\documentclass[pdflatex,sn-apa]{sn-jnl}

\usepackage{graphicx}%
\usepackage{multirow}%
\usepackage{amsmath,amssymb,amsfonts}%
\usepackage{amsthm}%
\usepackage{mathrsfs}%
\usepackage[title]{appendix}%
\usepackage{xcolor}%
\usepackage{textcomp}%
\usepackage{manyfoot}%
\usepackage{booktabs}%
\usepackage{algorithm}%
\usepackage{algorithmicx}%
\usepackage{algpseudocode}%
\usepackage{listings}%

\theoremstyle{thmstyleone}%
\theoremstyle{thmstyletwo}%

\theoremstyle{thmstylethree}%

\def\orcidlogo{}
\begin{document}

\title[Interpretable recognition of cognitive distortions in natural language texts]{Interpretable recognition of cognitive distortions in natural language texts}


\author*[1]{\fnm{Anton} \sur{Kolonin}}\email{akolonin@gmail.com}

\author[1]{\fnm{Anna} \sur{Arinicheva}}
\equalcont{These authors contributed equally to this work.}

\affil*[1]{\orgdiv{The Artificial Intelligence Research Center}, \orgname{Novosibirsk State University}, \orgaddress{\street{Pirogova 1}, \city{Novosibirsk}, \postcode{630090}, \state{State}, \country{Russia}}}


\abstract{We propose a new approach to multi-label classification of natural language texts based on weighted structured patterns such as N-grams, taking into account the heterarchical relationships between them, applied to solve such a socially impactful problem as the automation of detection of specific cognitive distortions in psychological care, relying on an interpretable, robust and transparent artificial intelligence model. The proposed recognition and learning algorithms significantly improve the current state of the art in this field. The improvement is tested on two publicly available datasets, with substantial improvements over literature-known $F1$ scores for the task, with optimal hyper-parameters determined, having the code and models available for future use by community.}




\maketitle

\section{Introduction}\label{sec:introduction}

\subsection{Significance}\label{sec:significance}

The prevalence of mental disorders has increased worldwide in recent decades, exerting growing pressure on society and healthcare systems \citep{Wu2023,UDUPA2023100013}. A large-scale multilingual analysis of natural language texts reports a substantial increase in cognitive distortions across cultures over the past thirty years \citep{Bollen2021}. Within cognitive behavioral therapy (CBT) \citep{Beck1963,beck1976cognitive}, such distortions are treated as key linguistic indicators of psychological disorders \citep{boyes2013common}.

This highlights the importance of providing robust tools to enable large-scale psychological help to improve well-being across societies, communities, smart cities, and enterprise employees. However, the availability of publicly available models and data for this purpose is quite limited, even for the English language. Therefore, we aim to develop technology that will provide transparent, interpretable, and cost-effective models for recognizing cognitive biases in natural language texts based on limited volumes of training data, improving on the state-of-the-art (SOTA) in this field and making the models available for public use. While we achieve this goal, as described below, only for English, the technology can be easily adapted to other languages, including low-resource languages, as well as other domains.

\subsection{Overview}\label{sec:overview}

Artificial intelligence (AI) and natural language processing (NLP) have been explored as tools to support mental health assessment and treatment \citep{Calvo2017NLE}, including automatic recognition of cognitive distortions in patient or client texts \citep{KOLONIN2022180}. However, several challenges continue to limit the practical use of such approaches.

First, the scarcity of openly available corpora containing clinical or counseling dialogues restricts the development of reliable models. In this setting, interpretable or at least explainable methods are required to expose the linguistic patterns underlying model predictions and to allow therapists to verify, revise, or reject automated conclusions. Interpretability further enables expert inspection and modification of models trained on limited or biased data.

Second, computational efficiency at inference time is critical. Due to regulatory constraints on storing or transmitting sensitive psychological data to third-party cloud services, cognitive distortion recognition should be performed locally on resource-constrained devices commonly used by practitioners, such as smartphones, tablets, or standard desktop computers without dedicated GPUs.

The social relevance of automated distortion recognition lies in its potential to support longitudinal monitoring of a client’s mental state. Changes in the frequency or intensity of cognitive distortions may provide objective indicators of deterioration or improvement. For example, \citet{https://doi.org/10.1002/cpp.70015} report that therapists correctly identified deterioration in only 1 out of 8 clients (12.5\%) based on self-reported symptom measures.

In this work, we propose an interpretable multi-class model for cognitive distortion recognition based on a structural representation of text patterns using heterarchical N-grams. N-grams are a relatively simple and well-established method; however, it is precisely this simplicity that ensures computational efficiency and interpretability of the resulting decisions, as well as transparency of the model itself, enabling auditing, revision, and expert-driven correction or extension. This makes N-gram-based approaches particularly suitable in settings with limited training data. In the context of cognitive distortion classification, the choice of an N-gram-based approach is further motivated by the fact that cognitive distortions manifest in language as explicit linguistic patterns \citep{Beck1963,beck1976cognitive}, and the only known open-source model in this domain is likewise based on N-grams \citep{Bollen2021}. We further introduce a computationally efficient recognition algorithm formulated as a convolution with an inverse kernel constrained by heterarchical relationships between patterns. The proposed training and recognition methods are evaluated on existing datasets, demonstrating improvements over previously reported SOTA $F1$ scores. All models and code are released as open source, providing a reproducible framework for interpretable, real-time cognitive distortion recognition using publicly available data.

\section{Related work}
\label{Related Work}

Natural language processing (NLP) methods have been extensively applied in related mental health domains. \citep{Zhang2022} review nearly 400 studies of NLP for mental illness detection and report an increase in such research. They find that deep learning models have largely supplanted traditional methods for identifying conditions like depression, anxiety, and PTSD, and they recommend developing novel and more interpretable architectures.

In clinical practice, NLP has been increasingly applied to support tasks such as patient screening, disease diagnosis, and clinical decision support \citep{Schopow-Nikolas-2023}. At a more specialized level, cognitive architectures have been proposed for use in psychotherapy to identify cognitive patterns and assist clinicians in detecting domain-specific constructs such as cognitive distortions \citep{KOLONIN2022180}.

Previous studies addressed cognitive distortion recognition using different formulations, ranging from binary detection of the presence of any distortion to multi-class recognition of specific distortion types. Most existing approaches rely on uninterpretable models and do not provide training data or open-source implementations. Interpretable models trained on publicly available datasets remain rare. Below, we review prior work grouped by model interpretability and data availability.

\subsection{Interpretable models with no data provided}
\label{Interpretable Models with No Data}

Several studies proposed interpretable approaches without releasing training or evaluation data. \citet{Simms2017} applied logistic regression with interpretable LIWC and RELIEF features for binary distortion detection, reporting an accuracy of 0.73. \citet{Shickel2019} used logistic regression with TF-IDF-based word features, achieving $F1=0.88$ for binary classification and $F1$ between 0.45 and 0.68 for multi-class recognition.

An interpretable model based on distortion-specific N-gram vocabularies was introduced in \citet{Bollen2021} to analyze long-term trends of cognitive distortions in large English, Spanish, and German text corpora. The study reported qualitative correlations between distortion prevalence and major negative societal events. These findings were later quantitatively supported by demonstrating causal relationships between cognitive distortions in social media and consumer behavior in cryptocurrency markets using the same model \citep{Kolonin2023_1}.

\subsection{Non-Interpretable models with no data provided}
\label{Non-Interpretable Models with No Data}

A number of works proposed uninterpretable models based on distributed representations, without publishing the underlying corpora. \citet{Zhao2017} applied CNN-based text classification using Word2Vec and CBOW embeddings for multi-class recognition but did not report $F1$ scores. \citet{Sochynskyi2021} used FastText for binary detection ($F1=0.71$) and a TF-IDF feature-based SVM for multi-class classification ($F1=0.23$). 

Transformer-based approaches include \citet{Tauscher2023}, which applied BERT to clinical dialogues and reported $F1=0.61$ for binary and multi-class tasks. Research and technology for detecting and analysing depression based on cognitive distortions among adolescent social media users with explainability option using BERT \citep{Wang2023} showed $F1$=0.78 in the multi-class recognition task. Multimodal models combining text with audio and visual features were explored in \citet{Vikram2023}, reporting $F1=0.74$ for binary detection. Using large language models (LLMs) to decipher cognitive distortions in patient-clinician mental health conversations using a multimodal LLM-based detection and reasoning framework underpinned by the LLAMA \citep{singh-etal-2024-deciphering} model, yielded $F1$=0.80 for the same binary task.

\subsection{Models with data provided}
\label{Models with Data}

The first publicly available dataset with multiple cognitive distortion labels derived from real therapist--client interactions was introduced by \citet{shreevastava-foltz-2021-dataset}. Using SVMs with BERT-based embeddings, \citet{shreevastava-foltz-2021-detecting} reported $F1=0.79$ for binary detection and $F1=0.30$ for multi-class recognition. Subsequent work introduced a synthetic dataset for binary distortion detection \citep{halil_2023} and applied RoBERTa-based models, achieving $F1=0.95$ on the binary task \citep{babacan2023creating}. A later study combined both datasets and extended labeling to multiple distortions \citep{halil_2024}, reporting macro $F1=0.32$ for non-synthetic dataset and $F1=0.76$ for semi-synthetic using RoBERTa \citep{article_1469178}. All these models are based on dense embeddings and lack interpretability.

The only study applying an interpretable approach to publicly available datasets is \citet{Arinicheva2025}. The authors employed an interpretable N-gram-based model originally proposed in \citet{Bollen2021} and introduced a pattern inference mechanism based on heterarchical relationships between N-grams, referred to as the \emph{priority on order} principle \citep{Raheman2022}. This approach achieved $F1=0.78$ for binary detection and $F1=0.21$ for multi-class recognition. According to this principle, higher-order N-grams take precedence over the lower-order N-grams they contain. For example, if the tetragram [“not", “a", “bad", “thing"] is matched, the contained bigram [“bad", “thing"] and unigram [“bad"] are ignored.

With the exception of the recognition code reported in \citet{Raheman2022}, none of the above studies released open-source implementations. The learning procedure for interpretable models applied in that work is described in \citet{Kolonin2022_2}, which is also referenced by \citet{Raheman2022}.

A general observation across prior work is that binary detection of cognitive distortions is substantially easier than multi-class recognition, resulting in consistently higher $F1$ scores. This is explained by both data sparsity and annotation ambiguity. For example, the bigram “I am" may be part of the trigram “I am responsible" (Personalization), “I am a" (Labeling), or a non-distorted utterance. Labeling inconsistencies may further limit learnability: in the dataset of \citet{halil_2024}, the sentence “I couldn't handle the workload, I'm bad at my job" appears twice, once labeled as Overgeneralization and once as Labeling. If such examples are split between training and test sets, accurate multi-class learning becomes technically infeasible.

\section{Methodology}
\label{Methodology}

We decided to focus our efforts on the more challenging problem of learning for interpretable recognition of specific cognitive distortions as a multi-classification task. Lacking open-source solutions beyond a basic N-gram model \citep{Bollen2021} and recognition concept \citep{Raheman2022}, we built the evaluation platform and learning pipeline entirely from scratch.

\subsection{Learning}
\label{Learning}  

The goal of the learning process is to create an interpretable N-gram model in the form of a heterarchy of patterns (N-grams) with different N. Individual segments of this heterarchy are associated with specific target categories of cognitive distortions, such as “All-or-nothing\_thinking", “Emotional\_Reasoning", “Fortune-telling", “Labeling", “Magnification", “Mental\_filter", “Mind\_Reading", “Overgeneralization", “Personalization", and “Should\_statements" aligned with the \citep{shreevastava-foltz-2021-dataset} dataset. Following \citep{Bollen2021}, each distortion is defined by a dictionary of N-grams acting as features. Training aims to select these distortion-specific features and their respective weights.

The learning of distortion-specific N-grams was conducted based on texts that were sentence-split via NLTK \citep{bird-loper-2004-nltk}, tokenized by spaces and punctuation, and stripped of punctuation. The N-gram space for selection was based on the maximum order of N-grams ($NM$), which ranged from 1 to 5.

For feature selection, we explored the possibility of using selection metrics ($SM$) from the list: $F$, $UF$, $FN$, $UFN$, $TF$-$IDF$, $FCR$, $CFR$, $MR$, $NLMI$, as defined below.

$|D|$ --- number of distortions in the entire corpus (10 in the case of our study).

$G_g$ --- frequency of occurrence of N-grams $g$ in the entire corpus.

$D_d$ --- frequency of distortion $d$ in the entire corpus defined as the number of texts labeled with it.

$G_d$ --- total number of N-grams $g$ associated with distortion $d$ over the entire corpus.

$UG_g$ --- number of unique N-grams $g$ across the entire corpus --- each N-gram is counted once for the entire corpus.

$D_g$ --- number of unique distortions associated with N-gram $g$ in the entire corpus (based on $UF_{gd}$).

$F_{gd}$ --- distortion-specific frequency of N-gram $g$ as the number of its associations with distortion $d$.

$UF_{gd}$ --- unique frequency of N-grams as the number of associations of distortion $d$ with N-gram $g$ or the unique number of associations of N-gram $g$ with distortion $d$ --- each N-gram is counted once in the text.

$\displaystyle FN_{gd} = \frac{F_{gd}}{G_g}$ --- N-gram frequency $g$ specific to distortion $d$, normalized by the N-gram frequency over the entire corpus.

$\displaystyle UFN_{gd} = \frac{UF_{gd}}{UG_g}$ --- unique N-gram frequency $g$ specific to distortion $d$, normalized by the unique N-gram frequency over the entire corpus.

$\displaystyle \mathit{TF\text{-}IDF}_{gd} = FN_{gd} \cdot log(\frac{|D|}{D_g})$ --- $TF$-$IDF$ metric, also used in earlier works \citep{Shickel2019,Sochynskyi2021}, is taken as the product of the normalized N-gram frequency of a distortion and the logarithm of the inverse N-gram frequency, i.e. how rare it is among all distortions.

$\displaystyle FCR_{gd} \frac{UF_{gd}}{\sum_d UF_{gd}}$ --- relevance of feature $g$ to category $d$, “feature-category relevance" according to \citet{Kolonin2022_2}.

$\displaystyle CFR_{gd} = \frac{UF_{gd}}{\sum_g UF_{gd}}$ --- relevance of category $d$ to feature $g$, “category-feature relevance" according to \citet{Kolonin2022_2}.

$\displaystyle MR_{gd} \frac{(UF_{gd})^2}{\bigl(\sum_d UF_{gd}\bigr) \cdot \bigl(\sum_g UF_{gd}\bigr)}$ --- “mutual relevance" between category $d$ and feature $g$, according to \citet{Kolonin2022_2}.

$\displaystyle NLMI_{gd} = \frac{(UF_{gd})^2}{D_d \cdot UG_g}$ --- non‑logarithmic mutual information between category $d$ and feature $g$.

To build weighted dictionaries for each distortion $d$, candidate features $g$ were scored using a normalized $SM$ defined above. An inclusion threshold $IT$ filtered out low-scoring features, and the qualifying features were saved into $d$-specific dictionaries.

$NM$, $SM$, and $IT$ were the learning hyper-parameters. Along with the detection threshold $DT$, which will be discussed later, they were used to find the optimal hyper-parameters in the corresponding 4-dimensional space.

\subsection{Recognition}
\label{Recognition}

\subsubsection{Weighted heterarchical pattern matching}

We implement weighted heterarchical pattern matching as an extended version of the recognition algorithm used in the earlier work of \citep{Raheman2022}, which can be seen in Algorithm~\ref{alg:priority_on_order}. It computes recognition scores $C$ as recognition metrics for each distortion based on the normalized number of N-grams in the text under evaluation $T$, given the presence of corresponding N-grams $g$ in distortion-specific dictionaries $\mathcal{F}$ learned by the model. In doing so, it takes into account the implicit heterarchical relationships between N-grams, ensuring that no N-gram with a smaller N that is part of a larger N-gram that is already matched is considered, according to the “priority on order" \citep{Raheman2022} principle. The latter is achieved by setting the mask $M$ to 0 in line 16 of the algorithm.

\begin{algorithm}[H]
\caption{Priority on order in recognition algorithm}
\label{alg:priority_on_order}
\begin{algorithmic}
\Require Input text $T$, cognitive distortions dictionaries $\mathcal{F}_1, \dots, \mathcal{F}_k$ with N-grams up to $NM$
\Ensure Normalized metric scores $\bar{C}_1, \dots, \bar{C}_k$
\State Tokenize $T$ into sequence $S = [s_1, \dots, s_l]$; let $i$ denote the token position in $S$, and $g$ the current N-gram starting at position $i$
\State Create mask $M = [1, \dots, 1]$ of length $l$
\State Initialize counts $C_j = 0$ for each metric $j = 1, \dots, k$
\For{$n = NM$ \textbf{to} $1$}
    \For{$i = 0$ \textbf{to} $l-n$}
        \If{$\sum_{t=0}^{n-1} M[i+t] = n$}
            \State $g \gets (s_i, \dots, s_{i+n-1})$
            \State $\mathit{found}\gets\text{false}$
            \For{$j = 1$ \textbf{to} $k$}
                \If{$g \in \mathcal{F}_j$}
                    \State $C_j \gets C_j + n \cdot H_{gj}$
                    \State $\mathit{found}\gets\text{true}$
                \EndIf
            \EndFor
            \If{$\mathit{found}$}
                \State $M[i:i+n]\gets 0$
            \EndIf
        \EndIf
    \EndFor
\EndFor
\For{$j = 1$ \textbf{to} $k$}
    \If{$LS$ enabled}
        \State $\bar{C}_j \gets \frac{1}{2} \log_{10}(1 + 100 \cdot C_j / l)$
    \Else
        \State $\bar{C}_j \gets C_j / l$
    \EndIf
\EndFor
\end{algorithmic}
\end{algorithm}

Algorithm~\ref{alg:priority_on_order} extends the original work by considering weighted matching in line 11 of the algorithm. $H_{gj}$ is the weight of feature $g$ for distortion $j$ in the model, determined as discussed in the previous section. In the unweighted version of the algorithm, $H_{gj} == 1.0$.

The final normalization of the $C$ scores can include logarithmic scaling if the $LS$ option is enabled in line 22 of the algorithm. In both logarithmic and non-logarithmic scaling cases, the scaling scheme ensures that the range of normalized metric values remains between 0 and 1. Our study of the algorithm showed that using this option does not affect the final training and recognition quality, as assessed by $F1$, since it only changes the optimal detection threshold intervals $DT$ due to the scaling of metric values. In our study, we used only logarithmic scaling.

For each metric $C$ computed for a particular distortion $j$ by the algorithm, scaled in the range from 0 to 100 percent, a decision on the presence or absence of a particular distortion in the text can be made based on the detection threshold $DT$.

\subsubsection{Recognition algorithm via convolution with inverse kernel}

For efficient computational implementation, the recognition algorithm described in Subsection~\ref{Recognition} can be redesigned using highly parallel vector operations, having texts and N-grams represented as vectors in a multidimensional space of tokens or words in a given lexicon, as follows. 
For simplicity, we omit here the demonstration of the masking implementation required to support the “priority on order" feature described above.

\noindent

\noindent\textbf{Text and Tokenization.}  
The original text \(T\) is tokenized into a sequence
\[
  S = (s_1, s_2, \dots, s_l),\quad l = |S|,\ \text{where}
\]
\hspace*{\parindent}$l$ — is the length of the input text.

\medskip
\noindent\textbf{N-gram dictionaries.}  
There are \(k\) dictionaries
\(\mathcal F_1,\dots,\mathcal F_k\), where
\[
\begin{aligned}
\mathcal F_j &= \{\,g^{(j)}_1,\dots,g^{(j)}_{M_j}\},\\
g^{(j)}_m &= (w_1,\dots,w_n),\; 1 \le n \le NM.
\end{aligned}
\]
Indices: \(j=1,\dots,k\), \(m=1,\dots,M_j\), \(n=1,\dots,NM\), where

$j$ — dictionary index, $m$ — index of the N-gram in the $j$-th dictionary, $n$ — length of the N-gram.

\medskip
\noindent\textbf{Overlap Range.}  
An N-gram of length \(n\) in the input text can be found at positions
\[
  i = 1,2,\dots,l - n + 1.
\]

\medskip
\noindent\textbf{Text Representation as a Vector.}  
Let \(V\) be the set of unique tokens from~\(S \cup \bigcup_{i=1}^{k} \mathcal{F}_i\). Denote the token index of the input text by \(f\).  
For each \(v \in V\) and \(f = 1, \dots, l\):
\[
  X_v(f)=
  \begin{cases}
    1,&s_f=v,\\
    0,&\text{otherwise},
  \end{cases}
  \quad
  \mathbf{x}_f=\bigl(X_v(f)\bigr)_{v\in V}
\]
\[
  \mathbf X=(\mathbf x_1,\dots,\mathbf x_l).
\]

\medskip
\noindent\textbf{Representation of N-grams as Vectors.}  
For each \(m\)-th N-gram of the \(j\)-th dictionary \(g^{(j)}_m = (w_1, \dots, w_n) \in \mathcal{F}_j\), consisting of \(n\) tokens, where \(p = 1, \dots, n\) is the position of the token in the considered N-gram:
\[
  H_{g,v}(p)=
  \begin{cases}
    1,&w_p=v,\\
    0,&\text{otherwise},
  \end{cases}
  \quad
  \mathbf{h}_{g,p}=\bigl(H_{g,v}(p)\bigr)_{v\in V},
\]
\[
  \mathbf H_g=(\mathbf h_{g,1},\dots,\mathbf h_{g,n}).
\]

\medskip
\noindent\textbf{Convolution with the Inverse Kernel.}  
For each dictionary \(j\), N-gram \(g \in \mathcal{F}_j\), length \(n\), and position \(i = 1, \dots, l - n + 1\):
\[
\begin{aligned}
y_g(i)
  &= \sum_{p=1}^{n}
     \langle \mathbf x_{i+p-1}, \mathbf h_{g,p} \rangle \\
  &= \sum_{p=1}^{n}
     \sum_{v \in V} X_v(i+p-1)\, H_{g,v}(p).
\end{aligned}
\]

\medskip
\noindent\textbf{Complete Match Indicator.}  
\[
  I_g(i)=
  \begin{cases}
    1,&y_g(i)=n,\\
    0,&\text{otherwise},
  \end{cases}
  \quad
  I_{j,n}(i)=\max_{\substack{g\in\mathcal F_j\\|g|=n}}I_g(i).
\]

\medskip
\noindent\textbf{Total Counters.}
\[
C_j = \sum_{n=1}^{NM} n \sum_{i=1}^{l - n + 1} I_{j,n}(i),
\]

\medskip
\noindent\textbf{Normalization.}  
\[
  \bar C_j=
  \begin{cases}
    \displaystyle \frac{C_j}{l},\\[1em]
    \displaystyle \frac12\log_{10}\Bigl(1+100\,\frac{C_j}{l}\Bigr).
  \end{cases}
\]

The convolution with the inverse kernel is used because otherwise, using the classical formulation, we recognize N-grams in the opposite order. For example, if we want to find the N-gram (“bad", “thing"), the convolution with the direct kernel will look for (“thing", “bad"), and the convolution with the inverse kernel will look for (“bad", “thing").

\subsection{Experimental setup}
\label{Experimental Setup}

The experimental work in our study included two datasets \citep{shreevastava-foltz-2021-dataset,halil_gpt4_2024} labeled with 10 cognitive distortions. The first set, obtained from the real field, contained 2530 texts, and the second included them but with an additional 2000 synthetic records artificially generated in the previous work \citep{babacan2023creating,article_1469178}.

For each of the two datasets, we applied stratified cross-validation in three independent runs on the corresponding splits or folds. 20\% of data was withheld for test set. The remaining 80\% underwent an 80/20 train/validation split across 3 shifted runs to ensure non-overlapping validation sets. Every fifth text was selected for the validation set, and the remaining four for the training set. This was done with shifts (0, 1, 2) specific to each of the three runs, so that completely different texts were included in each of the three validation splits.

Three runs for each of the two datasets involved evaluating an unweighted version of the recognition algorithm using the baseline model provided by earlier studies \citep{Bollen2021,Raheman2022}, as well as unweighted and weighted versions of the algorithm using multiple models learned in a three-dimensional space of maximum N-gram lengths $NM$, feature selection metrics $SM$, and feature inclusion thresholds $IT$. The $NM$ values were 1, 2, 3, 4, and 5. The $SM$ metrics were $F$, $UF$, $FN$, $UFN$, $TF$-$IDF$, $FCR$, $CFR$, $MR$, and $NLMI$. The $IT$ values were 0, 10, 20, 30, 40, 50, 60, 70, 80, and 90 percent.

For each combination of three learning hyper-parameters ($NM$, $SM$, $IT$), in each run, the trained model was evaluated on the recognition task with all $DT$ hyper-parameter values of 10, 20, 30, 40, 50, 60, 70, 80, and 90 percent.

Models were assessed via the unweighted average $F1$  (macro) score across all 10 distortions to obtain an overall $F1$ score used to evaluate the quality of the model for the corresponding $DT$. These $F1$ scores were aggregated across three independent runs with different training and test sets to obtain the average, minimum, maximum, and mean percentage error (MPE) for $F1$, to evaluate the significance of our results. During the evaluation, the models were used with the detection threshold ($DT$) determined during validation, thus eliminating any possibility of implicit leakage of data and hyper-parameters.The decision to use $F1$ as the target score was based on the fact that most previous work referenced it, allowing us to better compare our results with SOTA.

The primary goal of the experiments for each of the two datasets was to find the best average $F1$ score for all individual distortions and for three independent runs, and to determine the optimal combination of the aforementioned hyper-parameters that affect this score. A secondary goal was to examine suboptimal $F1$ scores obtained with other hyper-parameter combinations, and to compare the scores and hyper-parameter performance across models.

After the optimal hyper-parameters (including $DT$) and corresponding models for each of the two datasets were determined using 80\% of the data allocated for training and validation, a final evaluation was performed on a held-out test set comprising the remaining 20\% of the data. This evaluation was conducted using each of the three models corresponding to the different splitting, thereby enabling the calculation of the MPE.

The computing infrastructure used to conduct the experiments consisted of two laptops: 1) MSI Raider GE77HX 12UGS notebook with 12th Gen Intel(R) Core(TM) i7-12800HX 2.00 GHz, 32.0 GB RAM, 23.9 GB GPU NVIDIA GeForce RTX 3070 Ti Laptop GPU; 2) MacBook Pro with 2.9 GHz 6-Core Intel Core i9, Radeon Pro 560X 4GB Intel UHD Graphics 630 1536 MB, 32 GB 2400 MHz DDR4. The software used was Python version 3.11.13. The total computational budget was about three months. The final execution time for the presented experiment using the latter devices for each of the two Python Jupyter notebooks for the different datasets was about 4 hours. The entire code is available at \url{https://github.com/aigents/pygents/tree/main/papers/distortions\_multi\_2025}.

\section{Results and discussion}
\label{Results and Discussion}

\subsection{Model learning results}
\label{Model learning results}

\begin{figure}[htbp]
\centering
\includegraphics[width=0.75\columnwidth]{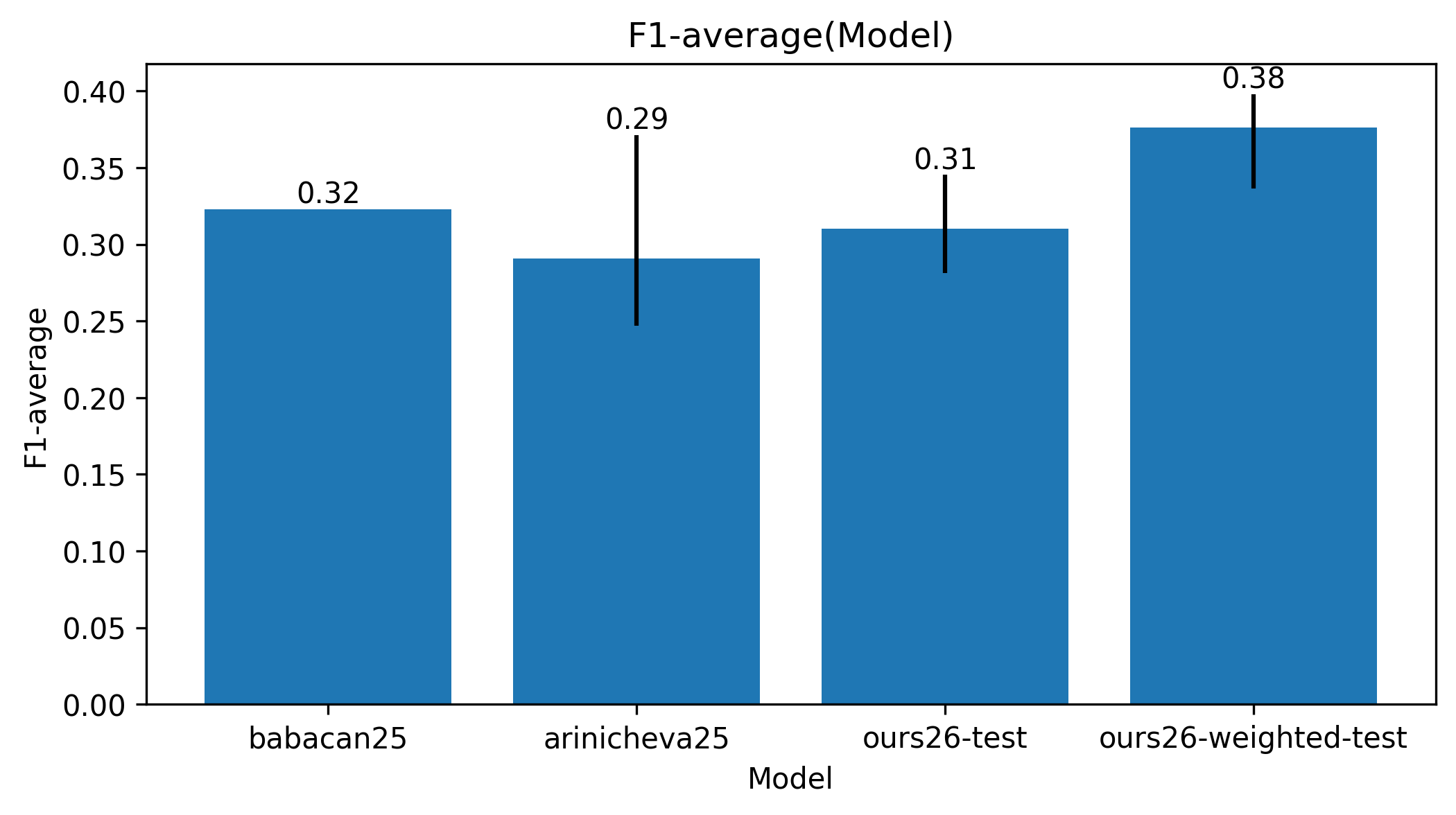}
\caption{Comparison of our $F1$ (macro) results on the real field data \citep{shreevastava-foltz-2021-dataset} using unweighted ($F1$=0.31) and weighted ($F1$=0.38) recognition with the baselines obtained in \citep{article_1469178} ($F1$=0.32) and based on the earlier model \citep{Arinicheva2025} ($F1$=0.29), with error bars for three independent train/test splits.}
\label{fig1_final_shreevastava}
\end{figure}

On the real field dataset \citep{shreevastava-foltz-2021-dataset}, our weighted model achieved a macro $F1$ of 0.38 (unweighted: 0.31), outperforming the baselines of 0.32 \citep{article_1469178} and 0.29 \citep{Arinicheva2025}, as shown in Figure~\ref{fig1_final_shreevastava}.

\begin{figure}[htbp]
\centering
\includegraphics[width=0.75\columnwidth]{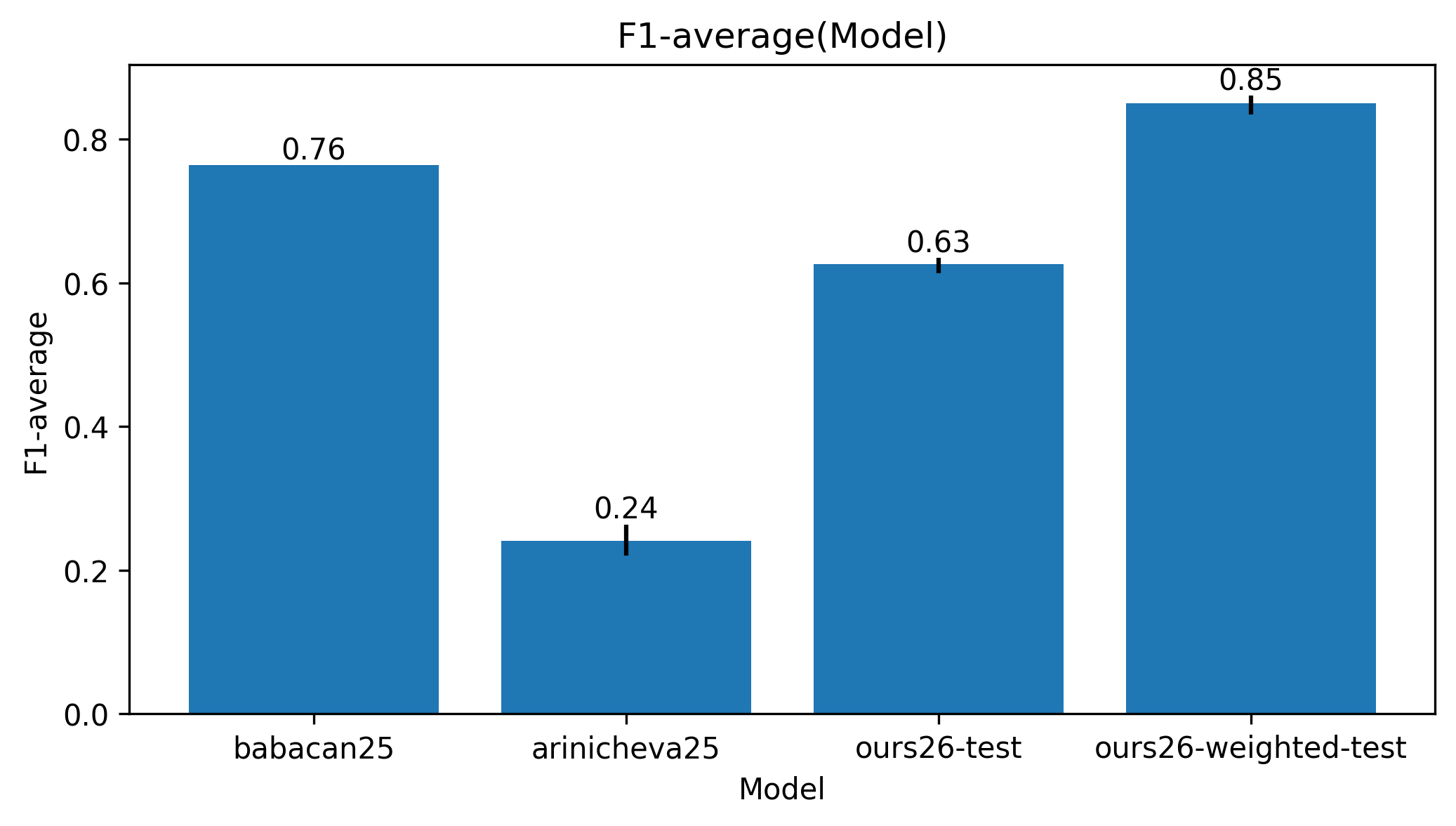}
\caption{Comparison of our $F1$ (macro) results on the combined semi-synthetic data \citep{halil_2024} using unweighted ($F1$=0.63) and weighted ($F1$=0.85) recognition versus the baselines obtained in \citep{article_1469178} ($F1$=0.76) and based on the earlier model \citep{Arinicheva2025} ($F1$=0.24), with error bars for three independent train/test splits.}
\label{fig2_final_babacan}
\end{figure}

On the semi-synthetic dataset \citep{halil_2024}, the weighted model reached $F1$=0.85 (unweighted: 0.63), exceeding prior baselines of 0.76 \citep{article_1469178} and 0.24 \citep{Arinicheva2025}, as shown in Figure~\ref{fig2_final_babacan}.

In experiments on the first dataset, the MPE was 7\%, and on the second, it was 1\%. In both cases, the MPE levels are smaller than the difference between our best results and earlier baseline values, confirming the significance of our results.

\begin{figure}[ht]
\centering
\includegraphics[width=0.85\columnwidth]{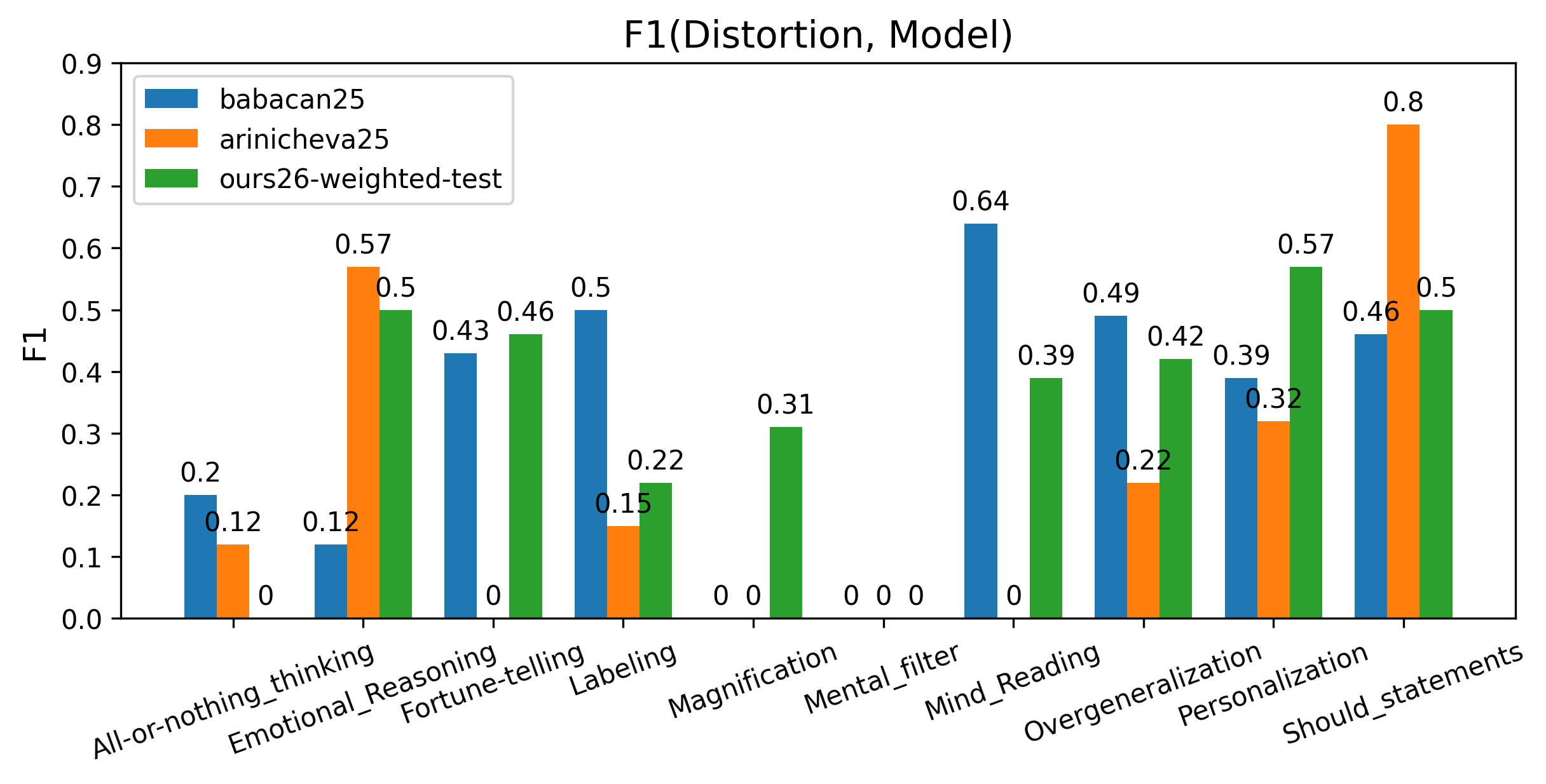}
\caption{Comparison based on the field dataset \citep{shreevastava-foltz-2021-dataset} with $F1$ scores obtained for certain distortions on the first split comparing our results (green) with the baseline presented in \citep{article_1469178} (blue) and based on the earlier model \citep{Arinicheva2025} (orange).}
\label{fig3_specific_shreevastava}
\end{figure}

Detailed analysis on the real field dataset \citep{shreevastava-foltz-2021-dataset} with optimal parameters ($NM$=2, $IT$=0, $SM$=$UFN$, $DT$=60, weighted) showed variance in metrics on Figure~\ref{fig3_specific_shreevastava}. While the overall $F1$ score for weighted recognition at $DT$=60 was 0.38, the $F1$ values for specific distortions ranged from 0.0 to 0.57 across a single split.

An in-depth analysis of recall and precision revealed that, while ideal recall (1.0) is achieved at $DT=50$, a precision of 1.0 is attained for half of the distortion types only when $DT \ge 70$. At the optimal level of $DT=60$, a completeness of 1.0 was recorded for five distortion types; for one type, it was 0.42; and for the remainder, it fell within the range of 0.58–0.78. At this specific $DT$ value, precision across all distortion types varied from 0.22 to 0.5. This indicates that achieving an acceptable level of recall presents no difficulty; however, issues typically arise specifically with precision.

Based on the obtained results, it can be concluded that our interpretable approach sets a new SOTA level in this field. Furthermore, the study confirms that weighted models possess a higher generalization capability, as they demonstrate significantly better results compared to unweighted models on the held-out test data for both sets.

\subsection{Hyper-parameters}
\label{Hyper-parameters}

For both datasets, the optimal maximum N-gram length was $NM$=2 (including unigrams and bigrams). Using $NM$=1 yielded poor quality, while $NM$>2 either decreased (the first dataset) or plateaued (the second dataset) the $F1$ score.

The choice of the best feature selection metric $SM$ to achieve optimal recognition quality turned out to be less stable. Below, we examine it based on the results of the first split.

For the first real field dataset, the best $SM$ metric was $UFN$ ($F1$=0.44-0.41), followed by $FN$ ($F1$=0.41-0.34) for both weighted and unweighted recognition. The $FCR$ metric ($F1$=0.37) was in third place for weighted recognition, and $NLMI$ ($F1$=0.32) for unweighted.

In the second semi-synthetic dataset, the best $SM$ metrics in the unweighted case turned out to be $MR$ and $NLMI$ ($F1$=0.89), followed by $FCR$, $FN$, and $UFN$ ($F1$=0.79). However, in the weighted case, conversely, the best metrics were $FCR$, $FN$, and $UFN$ ($F1$=0.86), followed by $TF$-$IDF$ ($F1$=0.84).

Summarizing the choice of the selection metric, $UFN$ is the most robust $SM$, having proven effective on real-world data. $IT$ and $DT$ inversely compensate for each other; their optimal balance depends strictly on the chosen $SM$.

For the first dataset ($SM$=$UFN$), optimal settings favored a lower inclusion and higher detection threshold: $IT$=10, $DT$=60 (unweighted) and $IT$=0, $DT$=60 (weighted). An expectedly lower $IT$ value is associated with lower overall match scores resulting from weighting.

For the second semi-synthetic dataset, in the case of the unweighted model, the $MR$ metric was chosen as $SM$, and the optimally balanced combination was $IT$=90, $DT$=40, however, in the case of weighted recognition, $SM$=$FN$, and the balanced combination was conversely $IT$=40, $DT$=70.

The general recommendation based on the results of our work is to use the model obtained from the first realistic dataset \citep{shreevastava-foltz-2021-dataset} with parameters $NM$=2, $IT$=0, $SM$=$UFN$, $DT$=60, together with the weighted version of the recognition algorithm.

\subsection{Comparison with Large Language Models}

For comparison, we evaluated our best weighted model ($F1 = 0.38$) on the first real field dataset against the baselines of \citet{Arinicheva2025} and \citet{article_1469178}, as well as local LLMs: LLAMA 3.2 (3B), QWEN 2 (7B), and QWEN 2.5 (7B). These out-of-the-box (without fine-tuning), open-source LLMs were self-hosted at a temperature of 0 to maximize stability and reproducibility.

For all LLMs, based on the list of distortions present in the training/test sets, the following prompt was specified: \textit{"You are professional psycho-therapist experienced in cognitive-behavioral therapy. You can label texts with none or some of cognitive distortions, represented by the following labels:\{str(list\_of\_distortions)\}. When labeling, return only JSON array in square brackets of strings in double quotes representing the labels. Label this text for presence or absence of any cognitive distortions from given list: \{text\}"}.

\begin{figure}[ht]
\centering
\includegraphics[width=0.75\columnwidth]{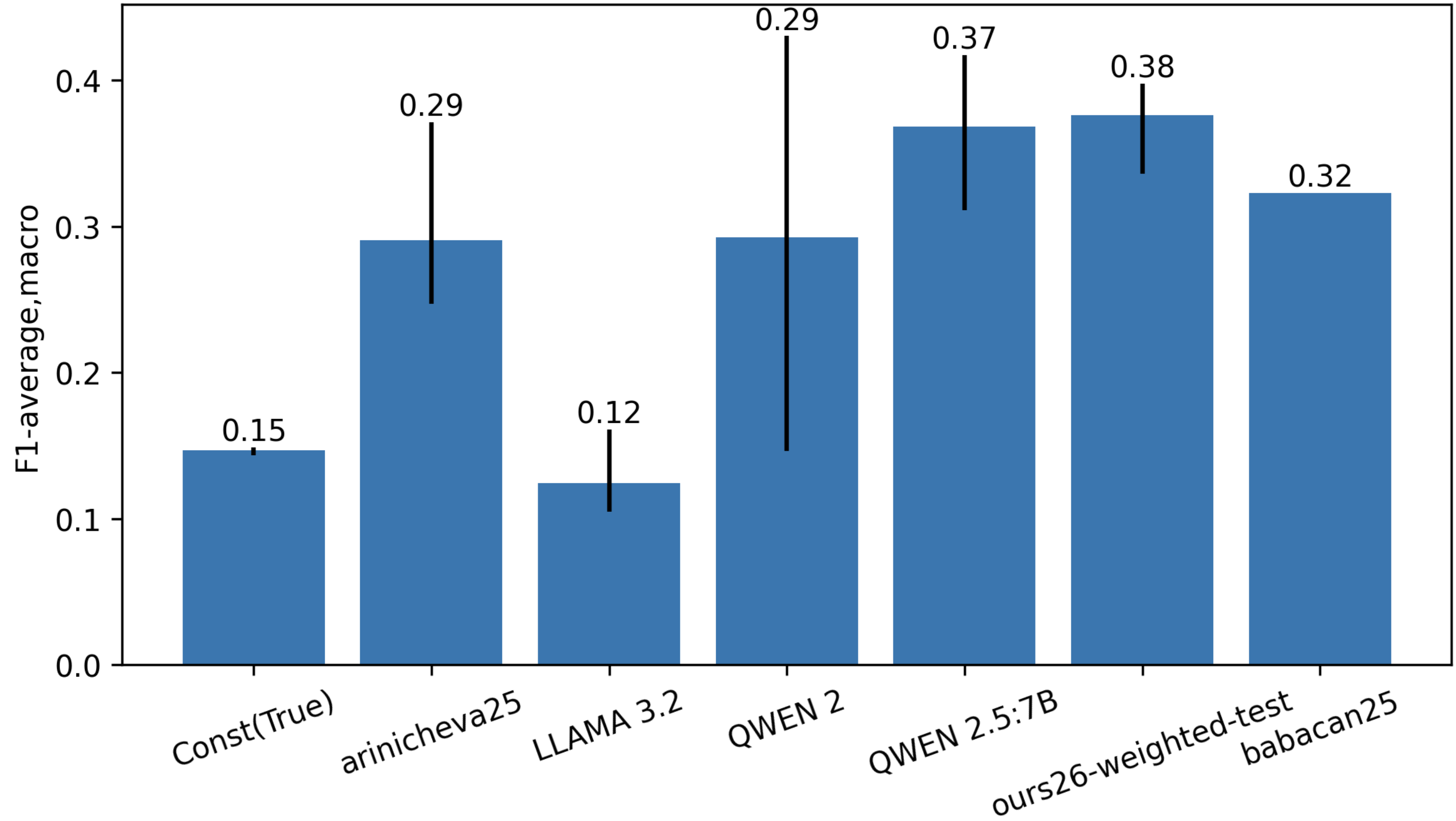}
\caption{Comparison of $F1$ (macro) scores on the real field data \citep{shreevastava-foltz-2021-dataset} between the $Const(True)$ baseline, the original model \citet{Arinicheva2025} (“arinicheva25"), three locally placed LLMs (“LLAMA 3.2", “QWEN 2", “QWEN 2.5:7B"), and our best weighted model (“ours26-weighted-test"), all based on three independent test splits, and the $F1$ based on \citet{article_1469178}: $F1$ (macro) computed by us from reported multi-class $F1$ scores (“babacan25"). Error bars correspond to the maximum and minimum $F1$ values across distortions.}
\label{fig4_f1macros_llm}
\end{figure}

\begin{figure}[ht]
\centering
\includegraphics[width=0.95\columnwidth]{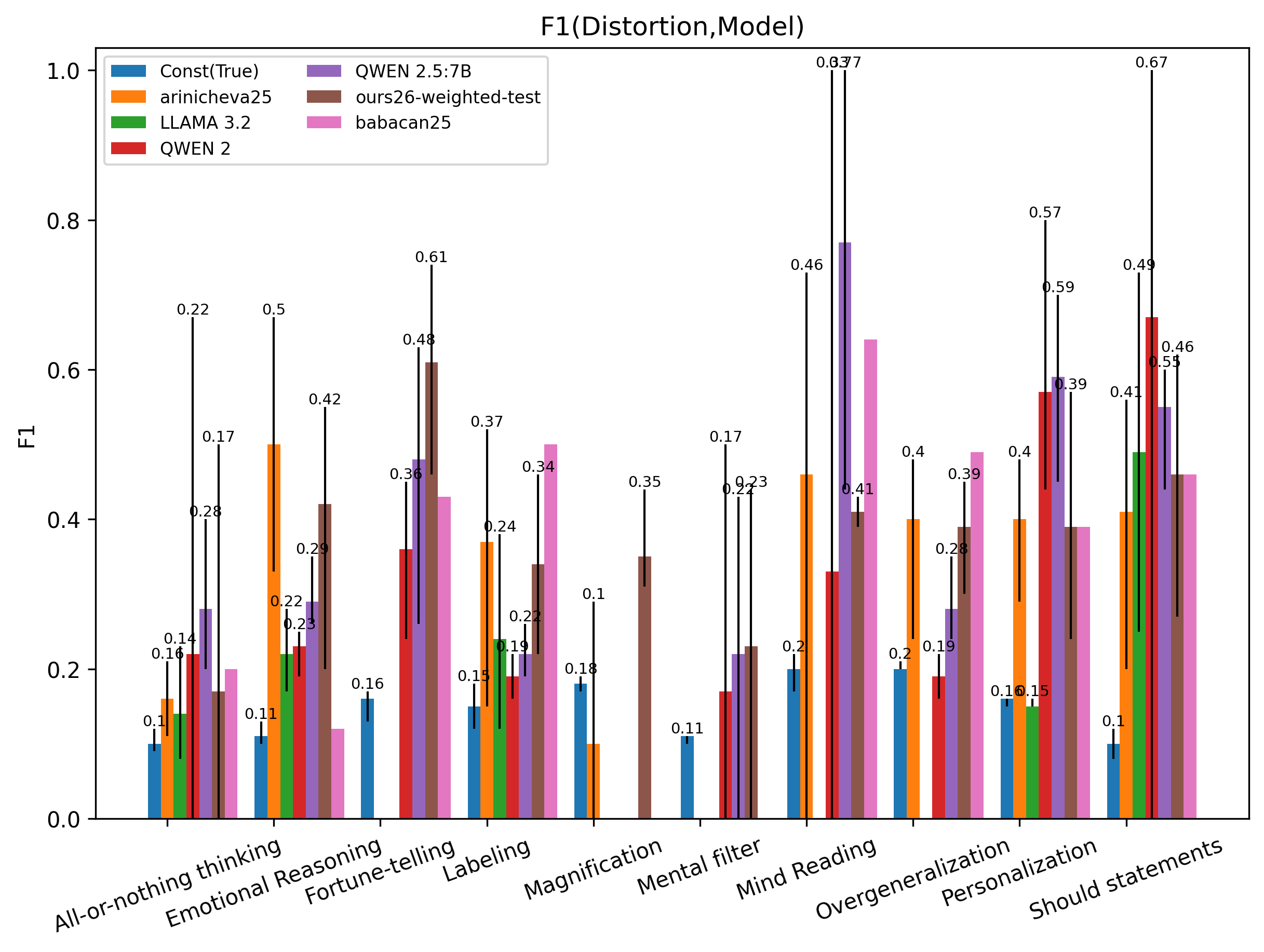}
\caption{Comparison of $F1$ scores on the real field dataset \citep{shreevastava-foltz-2021-dataset} for specific distortions between the $Const(True)$ baseline, the original model \citet{Arinicheva2025} (“arinicheva25"), three locally placed LLMs (“LLAMA 3.2", “QWEN 2", “QWEN 2.5:7B"), our best weighted model (“ours26-weighted-test"), and the scores reported in \citet{article_1469178} (“babacan25"). Error bars correspond to the maximum and minimum $F1$ scores obtained from three independent test splits.}
\label{fig5_f1multiclass_llm}
\end{figure}

\begin{figure}[ht]
\centering
\includegraphics[width=0.75\columnwidth]{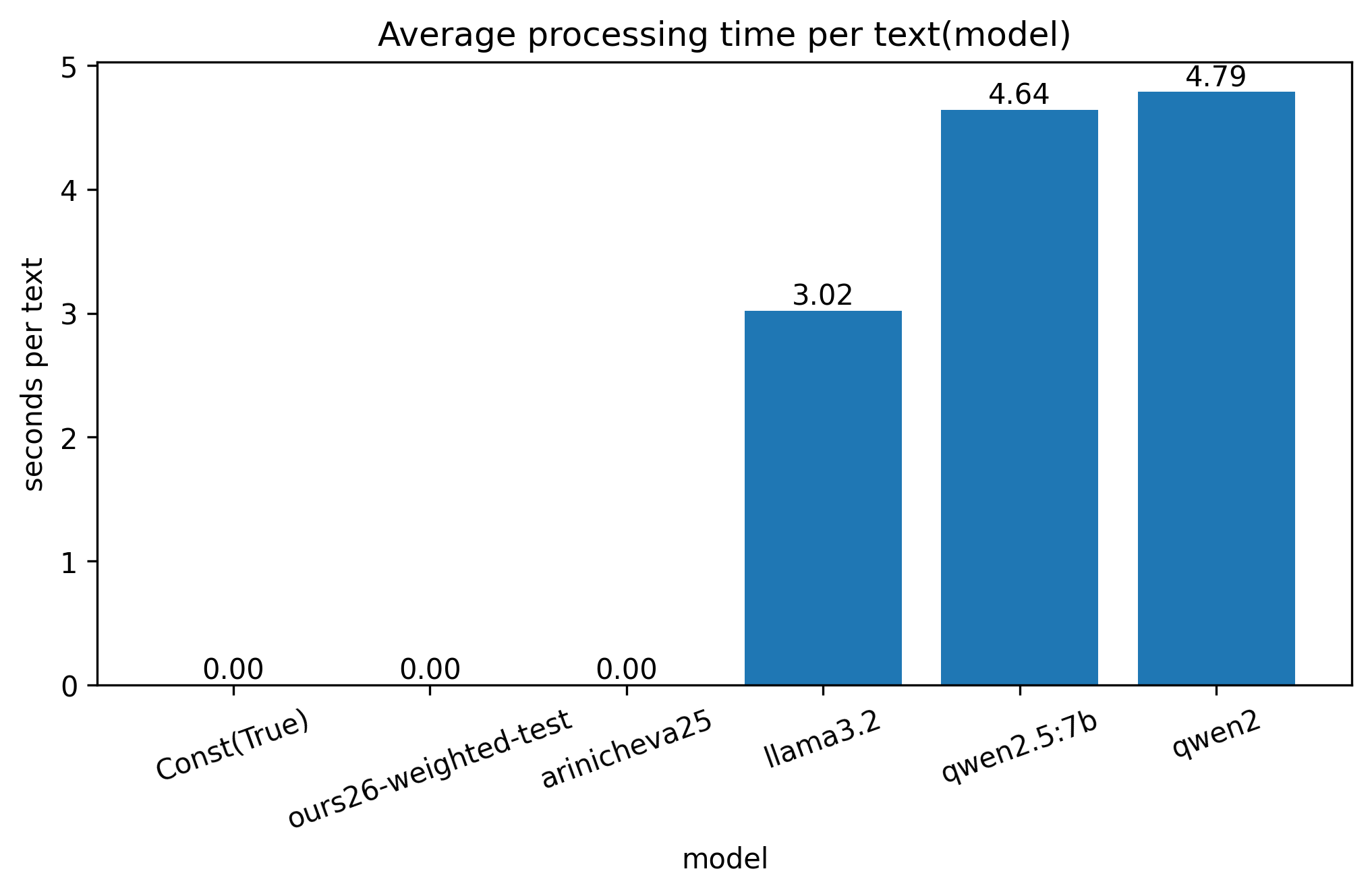}
\caption{Comparison of time costs for processing a single text from the real practical data \citep{shreevastava-foltz-2021-dataset} between the $Const(True)$ baseline, our best weighted model (“ours26-weighted-test"), the original model of \citet{Arinicheva2025} (“arinicheva25") and three locally deployed LLMs (“LLAMA 3.2", “QWEN 2", “QWEN 2.5:7B")}
\label{fig6_llm_time}
\end{figure}

In addition, the results were compared with an artificially created baseline using the $Const(True)$ model, which always detects all distortions in any texts, based on the assumption that any results worse than this level are meaningless.

Figure~\ref{fig4_f1macros_llm} shows average macro $F1$ scores evaluated across the same three independent test splits mentioned above, including error bars corresponding to MPE calculation needed to evaluate significance of the differences. 

LLAMA 3.2 performed near the $Const(True)$ baseline ($F1=0.12$), showing high instability (range 0.10–0.15). It was followed by \citet{Arinicheva2025} ($F1=0.29$, range 0.20–0.33), QWEN 2 ($F1=0.29$, range 0.16–0.44), and \citet{article_1469178} ($F1=0.32$). While QWEN 2.5:7B ($F1=0.37$, range 0.34–0.42) outperformed these baselines, our “ours26-weighted-test" model achieved the highest performance and stability ($F1=0.38$, range 0.34–0.40).

The maximum F1-macro performance achieved on the field dataset \citep{shreevastava-foltz-2021-dataset} can be further explained by discrepancies in the original data annotation reported by the authors of the dataset \citep{shreevastava-foltz-2021-dataset}: “On comparing the dominant distortion of about 730 data points encoded by two annotators, the Inter-Annotator Agreement (IAA) for specific type of distortion was 33.7\%. Considering the secondary distortion labels as well and computing a more relaxed agreement rate bumped the agreement to $\sim40\%$." Based on an analysis of manually and algorithmically assigned labels, the high variance in $F1$-scores observed across different data splits and distortion types can be plausibly attributed to the ambiguity inherently associated with the subjective annotation of field data: in such instances, the same text may be classified by different psychologists—or within different contexts—as belonging to one specific distortion type, or simultaneously to both.

The latter observation is supported by study on $F1$ and MPE across different distortions using different models as presented in Figure~\ref{fig5_f1multiclass_llm}, expending the $F1$ (macro) study shown in Figure~\ref{fig4_f1macros_llm}. The comparison was performed using the same three test splits discussed above, with independent $F1$ values evaluated for each model and distortion three times, so that the MPE could be calculated and error plots for the $F1$ values in Figure~\ref{fig5_f1multiclass_llm} could be drawn.

All that means that a corresponding limitation on highest possible $F1$ score should be expected for the field dataset \citep{shreevastava-foltz-2021-dataset} dataset, so our highest $F1=0.38$ on is close to the upper limit of what can be achieved using this particular dataset using traditional machine learning based on human-labeled training data.

At the same time, the average response time for LLMs was 4000 times longer than for our model as shown in \autoref{fig6_llm_time}. Consequently, the high computational costs, lack of interpretability, and comparable or lower $F1$ scores of LLMs render them impractical for this task.

A systematic evaluation of the F1 performance for cloud LLMs was not performed in this study due to privacy concerns discussed in section ~\ref{sec:introduction}, and because the response time performance analysis previously conducted in \cite{kolonin2026timeseriesaugmentedgeneration} showed that cloud LLMs perform even worse than on-premises models.

\subsection{Model interpretability}
\label{sec:interpretability}

\begin{figure}[ht]
\centering
\includegraphics[width=0.99\columnwidth]{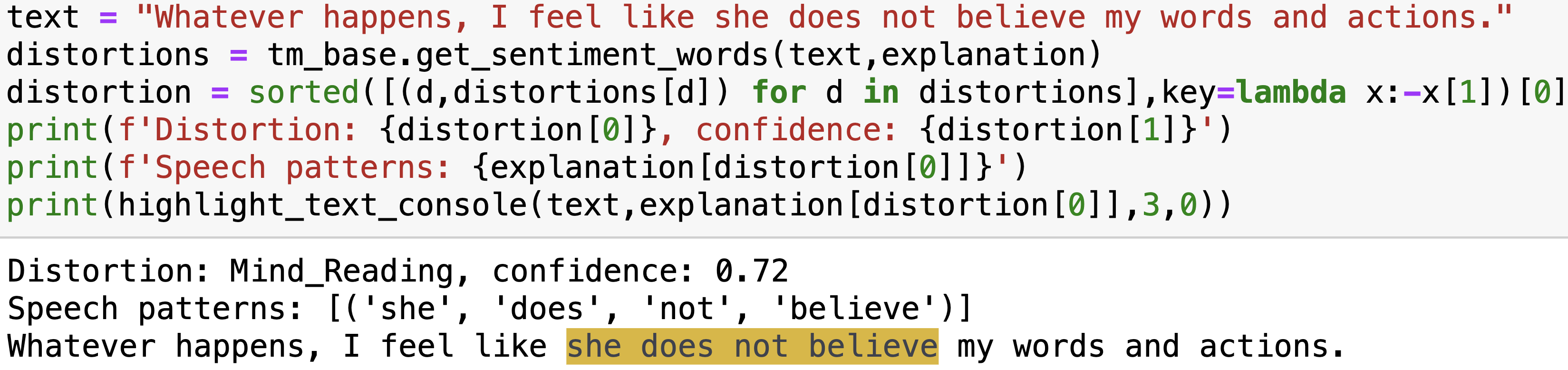}
\caption{This figure illustrates local interpretability, showing how the model explains its prediction for a single input text. Each highlighted N-gram contributes to the classification of a cognitive distortion, showing which specific linguistic patterns influenced the decision.}
\label{fig7_explainability}
\end{figure}

According to \citep{bassan2024localvsglobalinterpretability}, local interpretability characterizes the explanation of a model’s decision for a specific instance, whereas global interpretability reflects the general behavioral patterns of the model and the principles of its operation. We aimed to achieve both objectives: local interpretability, or explainability, as described in \citep{KOLONIN2022180}, is demonstrated in our approach in Figure~\ref{fig7_explainability}, while global interpretability is achieved by representing the model in the form of N-gram dictionaries following \citep{Bollen2021}.

As an example of global interpretability, we provide the most representative N-grams for the “Catastrophizing” distortion considered in our models, according to \citep{Bollen2021}: “will fail", “will go wrong", “will be impossible", “will not happen".

The possibility to view a model as an N-gram dictionary, which is a heterarchy of N-grams of varying order (N), allows an expert to validate or audit the model, remove erroneous entries introduced due to overfitting or mislabeling in the training corpus, and extend the model based on expert knowledge or external sources, as described in \autoref{cleaned}.

\subsection{Engineering a cleaned model for applications}
\label{cleaned}

Models obtained through cross-validation of the original field dataset \citep{shreevastava-foltz-2021-dataset} were examined to create a production model for practical use. An examination of the learned N-grams in the resulting vocabularies revealed a large number of “noisy" words and word combinations, apparently unrelated to cognitive distortions, which are features caused by overfitting, as expected given the small size of the training dataset.

First, to minimize the impact of the overfitting, we created a “joint" model based on the three best unweighted cross-validation models presented earlier in this paper, with an average F1-macro score of 0.46. The “joint" model included only N-grams present in all three independent models, significantly reducing the size of the resulting dictionaries for every distortion category, so this model could be considered “conservative." The model was evaluated on the same three test splits and yielded a F1-macro score of 0.68 (MPE 5\%). This value turned out to be expectedly high, which can be explained by data leakage due to the use of any of the test splits in the data used to learn some of the three split models.

\begin{figure}[htbp]
\centering
\includegraphics[width=0.75\columnwidth]{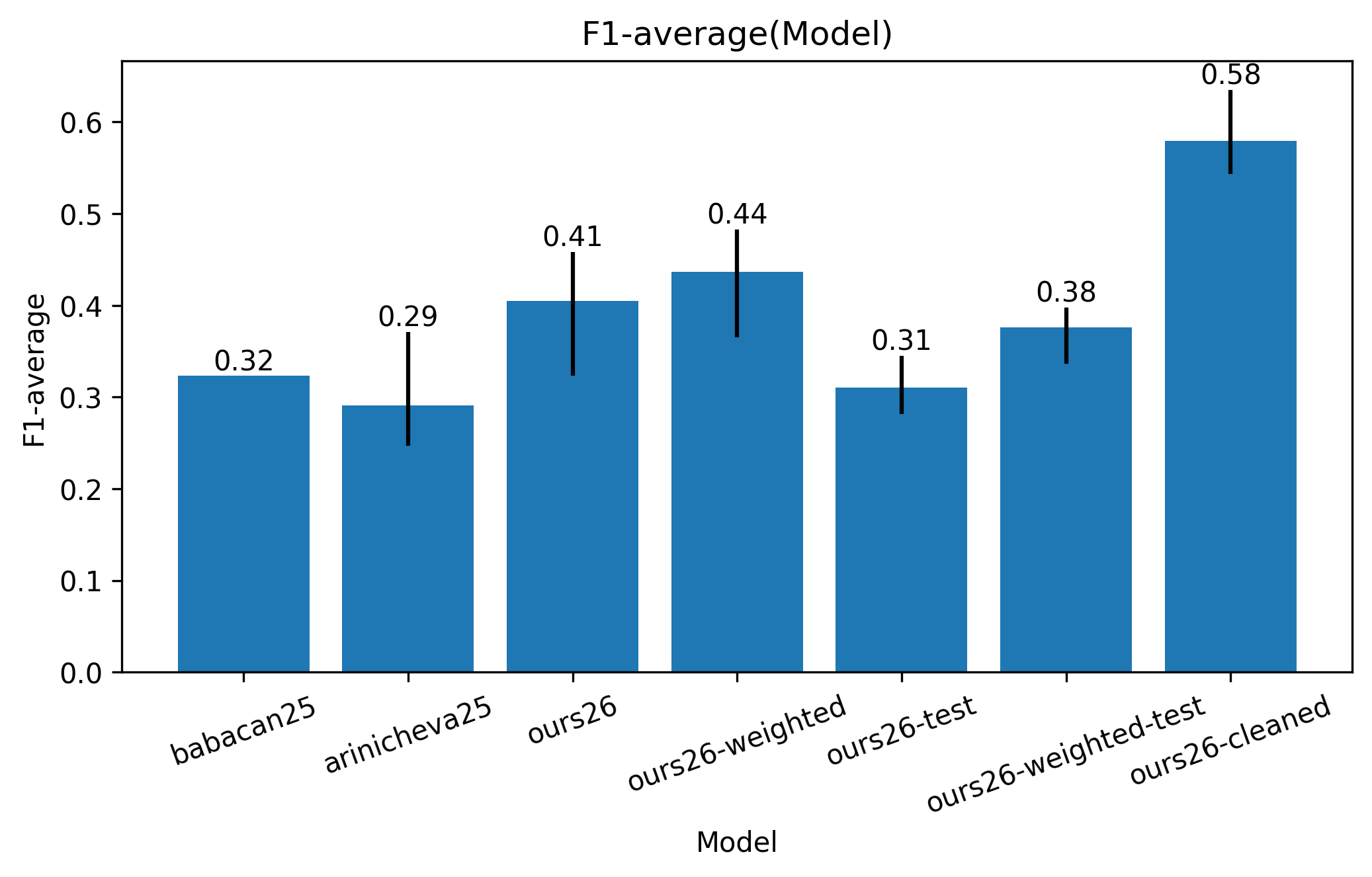}
\caption{Comparison of our $F1$ (macro) results on the real field data \citep{shreevastava-foltz-2021-dataset} using unweighted ($F1$=0.41 with no validation/test separation, $F1$=0.31 with separation) and weighted ($F1$=0.44 with no validation/test separation, $F1$=0.38 with separation) recognition with the baselines obtained in \citep{article_1469178} ($F1$=0.32), with the results based on the earlier model \citep{Arinicheva2025} ($F1$=0.29), with the results described in Section~\ref{Results and Discussion} and with a “cleaned" model ($F1$=0.58) with error bars for three independent train/test splits.}
\label{fig8_ours_cleaned}
\end{figure}

Next, the model N-gram dictionaries for each distortion were manually inspected to remove any “noisy" N-grams irrelevant to the psychological domain. The removal process was “optimistic", so we removed only those that were clearly off-topic (e.g., “the bus", “left leg", or “my mother"). However, the average proportion of removed content was approximately 90\% for each cognitive distortion dictionary.

Third, we enriched the remaining model using two external sources: N-gram dictionaries provided by \citet{Bollen2021} and N-gram lists generated by ChatGPT based on the prompt: \textit{“You are expert in computational linguistics, specialized in figures of speech and linguistic patterns indicating cognitive distortions known in cognitive-behavioral therapy. Which figures of speech or linguistics patterns can characterize cognitive distortion, known in cognitive behavioral therapy as “Mental Filtering"? Just provide plain list of all possible combinations of words”}. Using ChatGPT for this purpose, two queries with the same prompt were sent in two independent sessions; the different session results were combined and carefully checked for hallucinations, with redundant punctuation and formatting characters removed. According to \citet{Bollen2021}, we have created the model for 12 distortions rather than 10 used by \citet{shreevastava-foltz-2021-detecting}. 

Finally, the resulting N-gram lists for each distortion were sorted to remove duplicates, and each entry was manually augmented with multiple gender and tense forms, so that one “she lies" would become multiple forms, such as “he lies", “she lies", “they lie", “he lied", “she lied", “they lied", “he has lied", “she has lied", “they have lied".

The resulting joint, cleaned, enriched, and augmented model (we call it “cleaned") was evaluated using the same protocol described previously on three test splits, with results shown in \autoref{fig8_ours_cleaned}. It showed an $F1$-macro score of 0.58 (MPE 5\%) on the real field dataset \citep{shreevastava-foltz-2021-dataset}, which is higher than $F1$=0.29 achieved using an earlier interpretable model \citep{Arinicheva2025}, the best result for this dataset with $F1$=0.32 provided by \citet{article_1469178}, and our own results with weighted models ($F1$=0.44 with no test/validation separation, $F1$=0.38 with separation) reported above, and even above the pessimistic cap assumed by the dataset creators \cite{shreevastava-foltz-2021-detecting} based on labeling inconsistency discussed in \autoref{sec:data_limitations}.

Based on this, we can recommend for use in production the latest interpretable “cleaned" model, which provides the best quality based on the organic field dataset known in the field \citep{shreevastava-foltz-2021-dataset}. The model is available at \url{https://github.com/aigents/pygents/tree/main/data/models/distortions/shreevastava2021/joint-cleaned-10}. 

\section{Limitations}
\label{Limitations}

\subsection{Data limitations}
\label{sec:data_limitations}

The original field dataset \citep{shreevastava-foltz-2021-dataset} contains 2530 samples with no any information about the identity of the client involved in the dialogue with the psychologist whose utterances are annotated for cognitive distortions. The combined dataset \citep{halil_2024} includes both \citep{shreevastava-foltz-2021-dataset} and 2000 synthetic records from the \citep{halil_gpt4_2024} dataset, which were generated by ChatGPT-4 without using explicit “roles", meaning that, in fact, all the texts belong to a single author. As a result, when using the combined dataset \citep{halil_2024}, which combines the previous two, there is a possibility that texts from the same author may appear in both the training and test sets. This may result in the model learning stylistic features of the author's writing rather than actual indicators of cognitive distortions. 

The datasets described in Subsection~\ref{Models with Data} and used in this study are limited in size and cover only the English language, and the second dataset is partially synthetic. These datasets do not contain demographic, gender, or age information and may also fail to fully reflect the linguistic characteristics of real clinical interactions between clients and therapists. This may constrain the practical applicability of the models presented in this work; therefore, building more reliable models upon our approach, as well as models for other languages, will require additional and richer datasets, including corpora in multiple languages.

Both datasets are substantially imbalanced. For instance, the original field dataset \citep{shreevastava-foltz-2021-dataset} contains 2530 annotated sample texts, where 933 texts are annotated as “No Distortion". The remaining texts are annotated having one or more distortions (“Personalization": 202, “Labeling": 203, “Emotional Reasoning": 169, “Fortune-telling": 210, “Magnification": 245, “Mind Reading": 295, “All-or-nothing thinking": 126,  “Overgeneralization": 277, “Mental filter": 151, “Should statements": 135). Manual review of the labeled texts reveals multiple cases of inconsistent labeling, so that some texts with exactly the same meaning can be labeled with different distortions. That suggests that using this kind of data for a high-quality multi-label recognition model may be difficult, as discussed by \cite{shreevastava-foltz-2021-dataset}. 

\subsection{Baseline limitations}
\label{sec:baseline_limitations}

In both earlier works by \citet{shreevastava-foltz-2021-detecting} and  \citet{article_1469178} using the same datasets, it is not entirely clear how exactly the split into validation and test sets was performed in their experiments. Both papers state that the training and test sets were split in an 80/20 ratio. However, in the first paper \citep{shreevastava-foltz-2021-detecting}, a separate validation set is not mentioned, whereas in the second paper \citep{article_1469178}, both validation and test are mentioned in different places, but it is not specified whether the validation set was separated from the 80\% training set in some unspecified proportion or if the 20\% test set was also used for validation. To ensure adherence to the same protocol as in previous studies while preventing data leakage, in the cross-validation experiment described above—applied to both datasets—we first withheld 20\% of the original dataset as a test set; the remainder was then divided into 80\% training and 20\% validation sets (across three cross-validation splits).

Also, we found some inconsistency in the baseline SOTA results presented in \citet{article_1469178} regarding the $F1$ values. For example, Table 5 of the cited work reports an $F1$=0.45 (0.4467 in the paper) across label categories of the \citep{shreevastava-foltz-2021-dataset} dataset, but Table 2 reports individual $F1$ scores for specific distortions, while the $F1$ macro calculated from the latter numbers does not match the former number. In particular, if we exclude the “No Distortion" label according to our methodology, the calculated $F1$ (macro) for the \citep{shreevastava-foltz-2021-dataset} dataset is $F1$=0.32, which is lower than our best $F1$=0.38. One can only speculate that this may be due to the use of weighted macro accounts in \citet{article_1469178} or some other reasons not mentioned in that paper.

Furthermore, both papers reporting earlier baseline results and datasets used in our study \citep{article_1469178,shreevastava-foltz-2021-detecting} do not provide reproducible code, so it is not possible to verify the presented results with which we compare ours.

\subsection{Infrastructure limitations}

In the aforementioned comparative LLM experiment, which aimed to assess the feasibility of our approach on hardware typically available to psychologists and psychotherapists, we evaluated its performance under limited computing resources. To preserve data confidentiality in practical deployments, we intentionally considered only local (on-premise) deployment of LLMs, even though this may reduce achievable accuracy. Using an MSI Raider GE77HX (32.0 GB RAM, 23.9 GB GPU NVIDIA GeForce RTX 3070 Ti GPU) laptop as our test platform, we observed that our model achieved substantially faster processing times than locally run LLMs, demonstrating its suitability for real-time use in privacy-sensitive clinical settings.

\section{Conclusion}
\label{Conclusion}

We proposed an approach for building an interpretable model for recognizing cognitive distortions in psychological care texts. Our learning algorithm and enhanced recognition procedure were evaluated on two public datasets (real-world and semi-synthetic). Our model achieved new state-of-the-art $F1$ scores with reproducible code. The interpretable nature of the model helps identify distorted text segments, enabling experts to audit and extend the model to improve robustness.

Our improvement of the recognition algorithm by making it weighted allowed us to improve the recognition quality and make the model tolerant to the inclusion threshold.

For both datasets, we identified the optimal combinations of hyper-parameters for model learning and recognition and determined the optimal setting for further work. Based on the first real dataset, the “optimal" trained model, as well as a “cleaned" model ready for production use, are available at \url{https://github.com/aigents/pygents/tree/main/data/models/distortions/shreevastava2021/}.

In practice, our models may require expert review and correction, which is facilitated by their interpretability. Future research should focus on further automated post-training model cleaning and creating broader real-world datasets to minimize training and benchmark data sensitivity to labeling ambiguities.

To enhance interpretability and recall, another direction for future work would be to generalize learned patterns using formal grammars, domain ontologies, or sparse representations pre-trained on large unlabeled corpora.

Another direction for future research is to apply this technology to create interpretable models of cognitive distortion recognition in other languages (Chinese, Spanish, Russian, etc.) and other domains (such as emotion attribution and topic modeling).

\section{Ethics statements}
\label{sec:ethical-considerations}

\subsection{Potential risks}
\label{sec:risks}

The risks associated with technologies for cognitive distortion detection are similar to those discussed for sentiment analysis and related NLP applications \citep{Krishna2023,Kerstin2024}. The main risks considered in this study include the impact of errors, misuse or improper use, and potential violations of privacy, as discussed below.

\subsubsection{Impact of mistake}
\label{sec:mistake}

If we were developing non-interpretable model used for detection of cognitive distortions, the major risk would be its misuse so the false positive detections would effect in incorrect diagnostics of the cognitive distortions. However, since our model is based on interpretable and human-readable patterns, any user of the model can inspect it and have its decisions explained, so the risk can be taken under control. Moreover, a professional user can even adjust the model manually having the risk mitigated.

If the solution we provide is used to learn new models on insufficient or biased training data sets, or relying on the models that we present in this study blindly, without proper inspection, tuning and adjustment, false positive cognitive distortion diagnosis on behalf of psychologist or psychotherapist could result in inappropriate treatment. At the same time, while conventional non-interpretable solutions make this problem impossible to address, we make it possible to review, adjust and fine-tune the interpretable models manually by expert.

\subsubsection{Misuse}
\label{sec:misuse}

Any technology intended for psychological diagnostics may be misused. Professional users, such as psychologists or psychotherapists, may apply the system without sufficient validation or oversight. Non-professional users may also misuse the technology by placing excessive trust in automated outputs or by drawing non-professional conclusions from diagnostic results.

While the proposed approach may improve the efficiency and consistency of psychological diagnostics and could potentially support self-assessment, professional validation remains essential. Non-professional users should not rely on the system in isolation and should seek confirmation from qualified specialists before making decisions related to diagnosis or treatment.

Because the approach allows models to be trained on arbitrary text corpora, misuse may also occur if models are trained on inappropriate or invalid data and then applied without expert control. However, compared to non-interpretable methods, the proposed interpretable models allow professionals to assess model quality and suitability before deployment, which mitigates misuse risks when appropriate care is taken.

\subsubsection{Privacy violation}
\label{sec:privacy}

The use of the proposed technology by psychologists or psychotherapists based on informed consent from clients constitutes an ethically acceptable use case. Similarly, its application by governmental or commercial organizations to analyse publicly available online and social media content for monitoring general trends is considered legitimate.

In contrast, applying the method to proprietary or private textual data obtained in violation of privacy may increase harm by adding analytical value to improperly collected data. This risk is not specific to the proposed approach but applies to any form of automated processing of unlawfully obtained personal data and should be addressed through established legal, organizational, and technical safeguards.

\subsection{Scientific artifacts}
\label{sec:artifacts}

\subsubsection{Datasets}

The combined dataset \citep{halil_2024} including both \citep{shreevastava-foltz-2021-dataset} and extra synthetic records from the \citep{halil_gpt4_2024} dataset. The license is not specified in the dataset files or online metadata, however we contacted the authors \citep{article_1469178} and they confirmed that it is released under the MIT license, so our use of it can be considered fair. 

The original field dataset \citep{shreevastava-foltz-2021-dataset} contains the 2530 annotated samples of the patient's input annotated manually \citep{shreevastava-foltz-2021-detecting} and available online \citep{shreevastava-foltz-2021-dataset}. The license is not indicated in the dataset files or online metadata, however the data set is available online for four years and it was referenced in multiple latest publications \citep{shreevastava-foltz-2021-detecting, shreevastava-foltz-2021-dataset, babacan2023creating, halil_2023, article_1469178} so we treat possibility of its use as fair. 

Both datasets are hosted on machine learning platforms and include metadata indicating their intended use for research and machine learning purposes. Manual inspection of the datasets revealed no personally identifiable information or offensive content. As the datasets were anonymized in the original studies \citep{shreevastava-foltz-2021-detecting, shreevastava-foltz-2021-dataset}, no additional anonymization was required.

Although temporary datasets were created at runtime for learning and evaluation splits, no new dataset artifacts are released. Therefore, no additional licensing requirements apply beyond those governing the original datasets.

\subsubsection{Models}

The model files for cognitive distortions detection, along with the code we developed during the study and its release, are licensed under the MIT License, with no limitations on intended use, except for unlawful activities.

The model files used as our baseline models are manually created based on data published in public work by \citet{Bollen2021} in a format aligned with the design referenced in work by \citet{Raheman2022}, which references the model data under MIT license. Moreover, the same model files were used in subsequent study by \citet{Arinicheva2025} earlier.

The model files do not contain any offensive content or information that can be used for identification of individual people, because they are derived from the training datasets that have no such content or information either.

\subsection{Human annotators and participants}
\label{sec:human}

No human annotators, crowd-workers, or other participants, aside from the authors, were involved in this research. All training and evaluation data were obtained from previously published datasets, as described in Section~\ref{Models with Data}.

\subsection{Use of AI assistants}
\label{sec:assistants}

No AI assistants (such as ChatGPT or Copilot) were used in conducting the research, developing the code, or preparing the manuscript.

\section*{List of abbreviations}
\label{List of abbreviations}
\begin{itemize}
  \item[] \textbf{AI}: Artificial Intelligence
  \item[] \textbf{BERT}: Bidirectional Encoder Representations from Transformers
  \item[] \textbf{CBOW}: Continuous Bag-of-Words
  \item[] \textbf{CBT}: Cognitive Behavioral Therapy
  \item[] \textbf{CFR}: Category-Feature Relevance
  \item[] \textbf{CNN}: Convolutional Neural Network
  \item[] \textbf{DT}: Detection Threshold
  \item[] \textbf{F}: Frequency
  \item[] \textbf{FCR}: Feature-Category Relevance
  \item[] \textbf{FN}: Frequency Normalized
  \item[] \textbf{JSON}: JavaScript Object Notation
  \item[] \textbf{GPU}: Graphics Processing Unit
  \item[] \textbf{IT}: Inclusion Threshold
  \item[] \textbf{LIWC}: Linguistic Inquiry and Word Count
  \item[] \textbf{LLM}: Large Language Models
  \item[] \textbf{MR}: Mutual Relevance
  \item[] \textbf{MPE}: Mean Percentage Error
  \item[] \textbf{NLP}: Natural Language Processing
  \item[] \textbf{NLMI}: Non‑Logarithmic Mutual Information
  \item[] \textbf{NLTK}: Natural Language Toolkit
  \item[] \textbf{NM}: N-gram Maximum order 
  \item[] \textbf{SM}: Selection Metric
  \item[] \textbf{SOTA}: State-Of-The-Art 
  \item[] \textbf{SVM}: Support Vector Machine
  \item[] \textbf{TF-IDF}: Term Frequency-Inverse Document Frequency
  \item[] \textbf{UF}: Unique Frequency
  \item[] \textbf{UFN}: Unique Frequency Normalized
\end{itemize}

\section*{Declarations}

\subsection*{Availability of data and material}

The datasets analysed during the current study are available in the Kaggle repository \url{https://www.kaggle.com/datasets/sagarikashreevastava/cognitive-distortion-detetction-dataset/} \citep{shreevastava-foltz-2021-dataset} and in the Huggingface repository \url{https://huggingface.co/datasets/halilbabacan/combined\_synthetic\_cognitive\_distortions} \citep{halil_2024}.

The models generated during the current study are available in the GitHub repository, \url{https://github.com/aigents/pygents/tree/2026BI/data/models/distortions/shreevastava2021}.

\subsection*{Competing interests}

The authors declare that they have no competing interests.

\subsection*{Funding}

This work was supported by a grant for research centers, provided by the Ministry of Economic Development of the Russian Federation in accordance with the subsidy agreement with the Novosibirsk State University dated April 17, 2025 No. 139-15-2025-006: IGK 000000C313925P3S0002.

\subsection*{Authors' contributions}

Both authors AK and AA equally contributed to this work, read and approved the final manuscript.

\subsection*{Acknowledgements}

Not applicable.

\bibliography{sn-bibliography}

@misc{kolonin2026timeseriesaugmentedgeneration,
      title={Time Series Augmented Generation for Financial Applications}, 
      author={Anton Kolonin and Alexey Glushchenko and Evgeny Bochkov and Abhishek Saxena},
      year={2026},
      eprint={2604.19633},
      archivePrefix={arXiv},
      primaryClass={cs.AI},
      url={https://arxiv.org/abs/2604.19633}, 
}

@article{Wu2023,
  author = {Wu, Y. and Wang, L. and Tao, M. and Cao, H. and Yuan, H. and Ye, M. and Chen, X. and Wang, K. and Zhu, C.},
  title = {Changing trends in the global burden of mental disorders from 1990 to 2019 and predicted levels in 25 years},
  journal = {Epidemiology and Psychiatric Sciences},
  year = {2023},
  month = {November},
  volume = {32},
  pages = {e63},
  doi = {10.1017/S2045796023000756},
  pmid = {37933540},
  pmcid = {PMC10689059}
}

@article{UDUPA2023100013,
title = {Increases in poor mental health, mental distress, and depression symptoms among U.S. adults, 1993–2020},
journal = {Journal of Mood and Anxiety Disorders},
volume = {2},
pages = {100013},
year = {2023},
issn = {2950-0044},
doi = {https://doi.org/10.1016/j.xjmad.2023.100013},
url = {https://www.sciencedirect.com/science/article/pii/S2950004423000135},
author = {Nikhila S. Udupa and Jean M. Twenge and Cooper McAllister and Thomas E. Joiner}
}

@article{Bollen2021,
    author = {Johan Bollen and Marijn ten Thij and Fritz Breithaupt and Alexander T. J. Barron and Lauren A. Rutter and Lorenzo Lorenzo-Luaces and Marten Scheffer}, 
    title = {Historical Language Records Reveal a Surge of Cognitive Distortions in Recent Decades},
    journal = {PNAS},
    year = {2021},
    volume = {118},
    number = {30},
    pages = {e2102061118},
    doi = {10.1073/pnas.2102061118}
}

@article{Beck1963,
    author = {Aaron T. Beck},
    title = {Thinking and Depression. {I}. Idiosyncratic Content and Cognitive Distortions},
    journal = {Arch Gen Psychiatry},
    year = {1963},
    volume = {9(4)},
    pages = {324-333},
    doi = {10.1001/archpsyc.1963.01720160014002},
    PMID = {14045261}
}

@book{beck1976cognitive,
  author    = {Beck, Aaron T.},
  title     = {Cognitive Therapy and the Emotional Disorders},
  publisher = {International Universities Press, Inc.},
  address   = {Madison, CT},
  year      = {1976},
  month     = {apr},
  day       = {1},
  isbn      = {0823610055}
}

@misc{boyes2013common,
  author        = {Boyes, Alice},
  title         = {50 Common Cognitive Distortions},
  howpublished  = {Blog post, \emph{Psychology Today}, “In Practice” series},
  month         = jan,
  day           = {17},
  year          = {2013},
  note          = {Accessed: 2025-05-02},
  url           = {https://www.psychologytoday.com/ca/blog/in-practice/201301/50-common-cognitive-distortions},
}

@article{Calvo2017NLE,
  author  = {Rafael Calvo and David Milne and Sazzad Hussain and Helen Christensen},
  title   = {Natural language processing in mental health applications using non-clinical texts},
  journal = {Natural Language Engineering},
  volume  = {23},
  number  = {5},
  pages   = {649-685},
  year    = {2017},
  doi     = {10.1017/S1351324916000383},
  url     = {https://doi.org/10.1017/S1351324916000383},
}

@article{KOLONIN2022180,
title = {Cognitive Architecture for Decision-Making Based on Brain Principles Programming},
journal = {Procedia Computer Science},
volume = {213},
pages = {180-189},
year = {2022},
note = {2022 Annual International Conference on Brain-Inspired Cognitive Architectures for Artificial Intelligence: The 13th Annual Meeting of the BICA Society},
issn = {1877-0509},
doi = {https://doi.org/10.1016/j.procs.2022.11.054},
url = {https://www.sciencedirect.com/science/article/pii/S1877050922017458},
author = {Anton Kolonin and Andrey Kurpatov and Artem Molchanov and Gennadiy Averyanov}
}

@article{Raheman2022,
    author = {Ali Raheman and Anton Kolonin and Igors Fridkins and Ikram Ansari and Mukul Vishwas},
    title = {Social Media Sentiment Analysis for Cryptocurrency Market Prediction},
    journal = {arXiv:2204.10185},
    year = {2022},
    doi = {10.48550/arXiv.2204.10185}
}

@inproceedings{Kolonin2023_1,
    author = {Anton Kolonin and Ali Raheman and Mukul Vishwas and Ikram Ansari and Juan Pinzon and Alice Ho}, 
    title = {Causal Analysis of Generic Time Series Data Applied for Market Prediction},
    booktitle = {International Conference on Artificial General Intelligence, Lecture Notes in Computer Science (LNAI, volume 13539)},
    year = {2023},
    pages = {30-39},
    doi = {10.1007/978-3-031-19907-3_4}
}

@inproceedings{shreevastava-foltz-2021-detecting,
    title = "Detecting Cognitive Distortions from Patient-Therapist Interactions",
    author = "Shreevastava, Sagarika  and
      Foltz, Peter",
    editor = "Goharian, Nazli  and
      Resnik, Philip  and
      Yates, Andrew  and
      Ireland, Molly  and
      Niederhoffer, Kate  and
      Resnik, Rebecca",
    booktitle = "Proceedings of the Seventh Workshop on Computational Linguistics and Clinical Psychology: Improving Access",
    month = jun,
    year = "2021",
    address = "Online",
    publisher = "Association for Computational Linguistics",
    url = "https://aclanthology.org/2021.clpsych-1.17/",
    doi = "10.18653/v1/2021.clpsych-1.17",
    pages = "151-158"
}

@misc{shreevastava-foltz-2021-dataset,
	author       = {Sagarika Shreevastava},
        title        = {Cognitive distortion detection dataset (Version 1)},
	year         = {2021},
	url          = {https://www.kaggle.com/datasets/sagarikashreevastava/cognitive-distortion-detetction-dataset/},
	publisher    = {Kaggle},
        note = "Online dataset at Kaggle, visited 25-March-2025"
}

@inproceedings{Arinicheva2025,
    author = {Anna Arinicheva and Anton Kolonin},
    title = {Diagnosis of Cognitive Distortions in Public, Group, and Personal Text Communications},
    booktitle = {Advances in Neural Computation, Machine Learning, and Cognitive Research VIII},
    year = {2025},
    editor = {Redko, V. and Yudin, D. and Dunin-Barkowski, W. and Kryzhanovsky, B. and Tiumentsev, Y.},
    publisher = {Springer Nature Switzerland},
    volume = {1179},
    pages = {337-344},
    doi = {10.1007/978-3-031-80463-2_31}
}

@article{Kolonin2022_2,
  author = {Anton Kolonin},
  title = {High-Performance Automatic Categorization and Attribution of Inventory Catalogs},
  journal = {arXiv:2202.08965},
  year = {2022},
  doi = {10.48550/arXiv.2202.08965}
}

@misc{Kerstin2024,
      title={The Ethical Aspects of Integrating Sentiment and Emotion Analysis in Chatbots for Depression Intervention}, 
      author={Kerstin Denecke and Elia Gabarron},
      year={2024},
      journal = {Frontiers in Psychiatry},
      volume = {15},
      url={https://doi.org/10.3389/fpsyt.2024.1462083},
      doi={10.3389/fpsyt.2024.1462083}, 
      note = "Online at \url{https://doi.org/10.3389/fpsyt.2024.1462083}, visited 25-March-2025"
}

@article{Krishna2023,
    author = {Krishna Karoo and Vikas Chitte},
    title = {Ethical Considerations in Sentiment Analysis: Navigating the Complex Landscape},
    journal = {International Research Journal of Modernization in Engineering Technology and Science},
    volume = {05},
    issue = {11},
    issn = {2582-5208},
    year = {2023},
    url = {https://www.doi.org/10.56726/IRJMETS46811},
    doi = {10.56726/IRJMETS46811}
}

@article{article_1469178, 
    title={Creating a Clinical Psychology Dataset with Synthetic Data: Automatic Detection of Cognitive Distortions Classified with NLP}, 
    journal={Fırat Üniversitesi Mühendislik Bilimleri Dergisi}, 
    volume={37}, 
    pages={83–92}, 
    year={2025}, 
    DOI={10.35234/fumbd.1469178}, 
    author={Babacan, Hakkı Halil and Oğuz, Ramazan and Beyitoğlu, Yahya Kemal}, 
    number={1}, 
    publisher={Fırat Üniversitesi} }

@misc{babacan2023creating,
  author = {Babacan, Hakkı Halil and Beyitoğlu, Yahya and Oğuz, Ramazan},
  title = {Creating a Clinical Psychology Dataset with Synthetic Data: Automatic Detection of Cognitive Distortions Classified with NLP},
  year = {2023},
  journal = {SSRN},
  note = {Available at SSRN: url{https://ssrn.com/abstract=4582307} or url{http://dx.doi.org/10.2139/ssrn.4582307}},
  doi = {10.2139/ssrn.4582307}
}

@misc{halil_2023,
	author       = {Babacan},
    title        = {autotrain-data-cognitive\_distortions (Revision 4bc1d87)},
	year         = {2023},
	url          = {https://huggingface.co/datasets/halilbabacan/autotrain-data-cognitive\_distortions},
	doi          = {10.57967/hf/1002},
	publisher    = {Hugging Face},
    note = {Online dataset at Hugging Face, visited 25-March-2025}
}

@misc{halil_2024,
	author       = {Babacan},
	title        = {combined\_synthetic\_cognitive\_distortions (Revision 9995a75)},
	year         = 2024,
	url          = {https://huggingface.co/datasets/halilbabacan/combined\_synthetic\_cognitive\_distortions},
	doi          = {10.57967/hf/2857},
	publisher    = {Hugging Face},
    note = {Online dataset at Hugging Face, visited 25-March-2025}
}

@misc{halil_gpt4_2024,
	author       = { Halil },
	title        = { cognitive\_distortions\_gpt4 (Revision 5fd0943) },
	year         = 2024,
	url          = { https://huggingface.co/datasets/halilbabacan/cognitive\_distortions\_gpt4 },
	doi          = { 10.57967/hf/2858 },
	publisher    = { Hugging Face }
}

@INPROCEEDINGS {Simms2017,
author = {Taetem Simms and Clayton Ramstedt and Megan Rich and Michael Richards and Tony R. Martinez and Christophe G. Giraud-Carrier},
booktitle = { 2017 IEEE International Conference on Healthcare Informatics (ICHI) },
title = { Detecting Cognitive Distortions Through Machine Learning Text Analytics },
year = {2017},
volume = {},
ISSN = {},
pages = {508-512},
doi = {10.1109/ICHI.2017.39},
url = {https://doi.ieeecomputersociety.org/10.1109/ICHI.2017.39},
publisher = {IEEE Computer Society},
address = {Los Alamitos, CA, USA},
month =Aug}

@article{Shickel2019,
  title={Automatic Detection and Classification of Cognitive Distortions in Mental Health Text},
  author={Benjamin Shickel and Scott Siegel and Martin Heesacker and Sherry Benton and Parisa Rashidi},
  journal={2020 IEEE 20th International Conference on Bioinformatics and Bioengineering (BIBE)},
  year={2019},
  pages={275-280},
  url={https://api.semanticscholar.org/CorpusID:202583676}
}

@article{Zhao2017,
  author = {Xuejiao Zhao and Chunyan Miao and Zhenchang Xing},
  title = {Identifying Cognitive Distortion by Convolutional Neural Network Based Text Classification},
  journal = {International Journal of Information Technology},
  volume  = {23},
  number = {1},
  pages = {1-12},
  year = {2017},
  url = {https://hdl.handle.net/10356/89482},
  note = {© 2017 Singapore Computer Society. Author version, accepted for publication.}
}

@mastersthesis{Sochynskyi2021,
  author = {Stanislav Sochynskyi},
  title = {Automated Cognitive Distortion Detection and Classification of Reddit Posts Using Machine Learning},
  school = {University of Tartu},
  year = {2021},
  type = {Master's thesis},
  url = {https://hdl.handle.net/10062/93136},
  note = {Chair of Natural Language Processing, Supervisor: Kairit Sirts, PhD}
}

@article{Tauscher2023,
  author = {Justin S Tauscher and Kevin Lybarger and Xiruo Ding and Ayesha Chander and William J Hudenko and Trevor Cohen and Dror Ben-Zeev},
  title = {Automated Detection of Cognitive Distortions in Text Exchanges Between Clinicians and People With Serious Mental Illness},
  journal = {Psychiatric Services},
  year = {2023},
  volume = {74},
  number = {4},
  pages = {407-410},
  doi = {10.1176/appi.ps.202100692},
  pmid = {36164769},
  month = {April},
  note = {Epub 2022 Sep 27}
}

@ARTICLE{Wang2023,
    AUTHOR = {Bichen Wang and Yanyan Zhao and Xin Lu and Bing Qin},
    TITLE = {Cognitive distortion based explainable depression detection and analysis technologies for the adolescent internet users on social media},
    JOURNAL={Frontiers in Public Health},
    VOLUME={Volume 10 - 2022},
    YEAR={2023},
    URL={https://www.frontiersin.org/journals/public-health/articles/10.3389/fpubh.2022.1045777},
    DOI={10.3389/fpubh.2022.1045777},
    ISSN={2296-2565}}

@InProceedings{Vikram2023,
author="Gopendra Singh and Soumitra Ghosh and Asif Ekbal and Pushpak Bhattacharyya",
editor="Kamps, Jaap
and Goeuriot, Lorraine
and Crestani, Fabio
and Maistro, Maria
and Joho, Hideo
and Davis, Brian
and Gurrin, Cathal
and Kruschwitz, Udo
and Caputo, Annalina",
title="DeCoDE: Detection of Cognitive Distortion and Emotion Cause Extraction in Clinical Conversations",
booktitle="Advances in Information Retrieval",
year="2023",
publisher="Springer Nature Switzerland",
address="Cham",
pages="156-171",
isbn="978-3-031-28238-6"
}

@inproceedings{singh-etal-2024-deciphering,
    title = "Deciphering Cognitive Distortions in Patient-Doctor Mental Health Conversations: A Multimodal {LLM}-Based Detection and Reasoning Framework",
    author = "Gopendra Vikram Singh and Sai Vardhan Vemulapalli and Mauajama Firdaus and Asif Ekbal",
    editor = "Al-Onaizan, Yaser  and
      Bansal, Mohit  and
      Chen, Yun-Nung",
    booktitle = "Proceedings of the 2024 Conference on Empirical Methods in Natural Language Processing",
    month = nov,
    year = "2024",
    address = "Miami, Florida, USA",
    publisher = "Association for Computational Linguistics",
    url = "https://aclanthology.org/2024.emnlp-main.1256/",
    doi = "10.18653/v1/2024.emnlp-main.1256",
    pages = "22546--22570"
}

@inproceedings{bird-loper-2004-nltk,
    title = "{NLTK}: The Natural Language Toolkit",
    author = "Bird, Steven  and
      Loper, Edward",
    booktitle = "Proceedings of the {ACL} Interactive Poster and Demonstration Sessions",
    month = jul,
    year = "2004",
    address = "Barcelona, Spain",
    publisher = "Association for Computational Linguistics",
    url = "https://aclanthology.org/P04-3031/",
    pages = "214--217"
}

@article{https://doi.org/10.1002/cpp.70015,
author = {Østergård, Ole Karkov and Grønnebæk, Lasse and Nilsson, Kristine Kahr},
title = {Do Therapists Know When Their Clients Deteriorate? An Investigation of Therapists' Ability to Estimate and Predict Client Change During and After Psychotherapy},
journal = {Clinical Psychology \& Psychotherapy},
volume = {31},
number = {6},
pages = {e70015},
doi = {https://doi.org/10.1002/cpp.70015},
url = {https://onlinelibrary.wiley.com/doi/abs/10.1002/cpp.70015},
eprint = {https://onlinelibrary.wiley.com/doi/pdf/10.1002/cpp.70015},
note = {e70015 CPP-4416.R1},
year = {2024}
}

@Article{Zhang2022,
author="Zhang, Tianlin
and Schoene, Annika M.
and Ji, Shaoxiong
and Ananiadou, Sophia",
title="Natural language processing applied to mental illness detection: a narrative review.",
journal="npj Digital Medicine",
year="2022",
volume="5",
doi="10.1038/s41746-022-00589-7",
url="https://doi.org/10.1038/s41746-022-00589-7"
}

@Article{Schopow-Nikolas-2023,
author="Schopow, Nikolas
and Osterhoff, Georg
and Baur, David",
title="Applications of the Natural Language Processing Tool ChatGPT in Clinical Practice: Comparative Study and Augmented Systematic Review",
journal="JMIR Med Inform",
year="2023",
month="Nov",
day="28",
volume="11",
pages="e48933",
issn="2291-9694",
doi="10.2196/48933",
url="https://doi.org/10.2196/48933",
}

@misc{bassan2024localvsglobalinterpretability,
      title={Local vs. Global Interpretability: A Computational Complexity Perspective}, 
      author={Shahaf Bassan and Guy Amir and Guy Katz},
      year={2024},
      eprint={2406.02981},
      archivePrefix={arXiv},
      primaryClass={cs.LG},
      url={https://arxiv.org/abs/2406.02981}, 
}

\end{document}